\documentclass[letterpaper, 10 pt, conference]{ieeeconf}  

\IEEEoverridecommandlockouts                              

\usepackage{graphicx}
\usepackage{amsmath} 
\usepackage{amssymb}  
\usepackage{flushend}

\newcommand{\Prob} {\ensuremath \mathbf{P}  }

\newcommand{\Indi} {\ensuremath \mathbf{1} }
\newcommand{\Hx} {\ensuremath \mathbf{H}  }

\graphicspath{{figs/}}

\title{\LARGE \bf
Anomaly Detection in Unstructured Environments using Bayesian Nonparametric Scene Modeling
}

\author{Yogesh Girdhar $^{1}$, Walter Cho$^{2}$, Matthew Campbell$^{3}$, Jesus Pineda $^{1}$, Elizabeth Clarke$^{3}$, and Hanumant Singh$^{1}$
\thanks{*This work was supported by the Postdoctoral Scholar Program at the Woods Hole Oceanographic Institution, with funding provided by the Devonshire Foundation and the J. Seward Johnson Fund, and  FQRTN Postdoctoral Fellowship.}
\thanks{$^{1}$Authors are at Woods Hole Oceanographic Institution, Woods Hole, MA 02543, USA.
        {\tt\small \{ygirdhar,hsingh,jpineda\}@whoi.edu}}%
\thanks{$^{2}$This author is with Point Loma Nazarene University, San Diego, CA 92106, USA. 
        {\tt\small waltercho@pointloma.edu}}%
\thanks{$^{3}$These authors are with NOAA, USA. 
        {\tt\small \{matthew.d.campbell,elizabeth.clarke\}@noaa.gov}}%
}

\begin{document}

\maketitle
\thispagestyle{empty}
\pagestyle{empty}

\begin{abstract}
This paper explores the use of a Bayesian non-parametric
topic modeling technique for the purpose of anomaly
detection in video data. We present results from two
experiments. The first experiment shows that the proposed
technique is automatically able characterize the underlying
terrain, and detect anomalous flora in image data collected
by an underwater robot. The second experiment shows that
the same technique can be used on images from a static
camera in a dynamic unstructured environment. In the second
dataset, consisting of video data from a static seafloor
camera capturing images of a busy coral reef, the proposed
technique was able to detect all three instances of an
underwater vehicle passing in front of the camera, amongst
many other observations of fishes, debris, lighting changes
due to surface waves, and benthic flora.

\end{abstract}

\section{Introduction}
Robots or stationary cameras when used for surveying and monitoring tasks collect large amounts of image data, which is often analyzed manually by human experts. At Woods Hole Oceanographic Institution (WHOI) and NOAA Fisheries for example, every year 1000s of hours of video is collected using AUVs and static cameras, and for every hour of video it current takes approximately 3-4 hours of manual processing time. Hence, there is a need for automated techniques that can speed up the analysis of such datasets by identify perplexing or anomalous observations. Through the use of such techniques we can focus the attention of the human expert on a small subset of the collected data that is most likely to contain relevant information. In this paper we explore the use of Bayesian non-parametric (BNP) topic modeling to detect and characterize such anomalies.

Compared to other kinds of sensor data, image data typically exists in millions of dimensions, corresponding to the number of pixels in the image, which makes it challenging to build an automatic anomaly detection technique.
Moreover, detecting anomalous events in a non-stationary unstructured environment, such as coral reefs is even more challenging due to its higher visual complexity, compared to urban scenes. 
 
Our proposed approach to dealing with the anomaly detection problem is to first use a Bayesian non-parametric scene understanding technique to build a model of the scene, and then using this model identify observations that are perplexing for the model.  BNP topic modeling techniques have been successful in building semantic models of the data that automatically grow in complexity with data complexity. We take advantage of these results, and extended our previous work on Realtime Online Spatiotemporal Topic Modeling (ROST) \cite{Girdhar2013IJRR} to incorporate Bayesian nonparametric priors. In this paper we refer to our resulting proposed scene modeling technique as BNP-ROST.



\section{Related Work}
\subsubsection{Topic Modeling}
Topic modeling techniques like Probabilistic Latent Semantic Analysis(PLSA) \cite{Hofmann:2001}, and Latent Dirichlet Allocation (LDA) \cite{Blei:2003,Griffiths:2004}, although originally developed for semantic analysis of text documents, they have been widely applied to other types of data such as images\cite{Bosch:2006, FeiFei:2005:CVPR, Wang2007}. The general idea behind topic modeling, as applied to image data is to describe each image in a dataset as a distribution over high level concepts, without having prior knowledge about what these concepts are.


Probabilistic Latent Semantic Analysis (PLSA)\cite{Hofmann:2001} models the probability of observing a word $w$ in a given document $m$ as:
\begin{eqnarray}
\Prob(w| d) = \sum_{k=1}^{K} \Prob(w | z=k) \Prob(z=k | d),
\end{eqnarray}
where $w$ takes a value between $1\dotsc V$, where $V$ is the vocabulary size;  $z$ is the hidden variable or topic label for $w$ that takes a value between $1\dotsc K$, where $K$ is the number of topics, and is much smaller than $V$; and $d$ is the document number, which can take a value between $1\dots M$, where $M$ is the total number of documents. The central idea is the introduction of a latent variable $z$, which models the underlying topic, or the context responsible for generating the word.  Each document $m$ in the given corpora is modeled using a distribution $\theta_m(k) = \Prob(z=k | d=m)$ over topics, and each topic is modeled using a distribution $\phi_k(v)= \Prob(w=v | z=k) $ over the set of vocabulary words. During the training phase,  these distributions are learned directly using an EM algorithm. 

The distribution of topics in a document gives us a low dimensional semantic description of the document, which can be used to compare it semantically with other documents. The problem with this approach is that since the dimensionality of the model is very large, a lot of training data is required. Moreover, it is easy to overtrain for a given data set.

Latent Dirichlet Allocation (LDA), proposed by Blei et al. \cite{Blei:2003} improves upon PLSA by placing a Dirichlet prior on $\theta$ and $\phi$, encouraging the distributions to be sparse, which has been shown to give semantically more relevant topics. Subsequently Griffiths et al. \cite{Griffiths:2004} proposed a Gibbs sampler to learn these distributions.

\subsubsection{Semantic Modeling of Image Data}
Topic modeling of images requires that the general idea of a textual word be replaced by visual words. One approach to generate these visual words from visual features is that described by Sivic et al.~\cite{Sivic:2006:videogoogle}. 
Given visual word representation of scenes with multiple objects, topic modeling has been used to discover objects in these images in an unsupervised manner. Bosch et al. \cite{Bosch:2006} used PLSA and a SIFT based~\cite{Lowe:IJCV:2004} visual vocabulary to model the content of images, and used a nearest neighbor classifier to classify the images. Fei-Fei et al.\cite{FeiFei:2005:CVPR} have demonstrated the use of LDA to provide an intermediate representation of images, which was then used to learn an image classifier over multiple categories. Instead of modeling the entire image as a document, Spatial LDA (SLDA) \cite{Wang2007} models a subset of words, close to each other in an image as a document, resulting in a better encoding of the spatial structure. The assignment of words to documents is not done \emph{a priori}, but is instead modeled as an additional hidden variable in the generative process. 

Summarizing benthic (sea floor) images is an especially difficult problem due to the general lack of order, symmetry, and orientation of the visual data. Steinberg et al. \cite{Steinberg2011} used a Gaussian mixture model to cluster benthic stereo images, while using a Variation Dirichlet Process~\cite{Kurihara2007} to automatically infer the number of clusters. Although this work did not use location information in the clustering process, the resulting cluster labels were shown to be spatially contiguous, indicating correctness. The computed labels were shown to outperform those obtained with spectral clustering and EM Gaussian mixture models, when compared with hand labeled ground truth data.

\subsubsection{Topic Modeling of Streaming Video Data}
BNP techniques have been previously used to characterize anomalous activities~\cite{Wang2007a, Hospedales2012}. In our own recent work \cite{Girdhar2013IJRR} we have described a realtime online topic modeling (ROST) technique that computes topic labels for observed visual words in a video while taking into account its spatial context in pixel space, and temporal context (frame count). ROST does this by generalizing the idea of a document to a spatiotemporal cell, and computing the topic label for the words in a cell in the context of its neighboring cells. In \cite{Girdhar2015a} we used ROST to identify interesting observations in a robot's view, and then used it to plan an adaptive path, which were shown to have higher information content than simple space filling paths.

\section{Bayesian Nonparametric (BNP) Scene Modeling}
 Given a sequence of images or other observations, we extract discrete features $w$ from these observations, each of which has corresponding spatial and temporal coordinates $(x,t)$. In case of a simple 2D video the spatial coordinates would just correspond to the pixel coordinates, however in presence of 3D data, the spatial coordinates can be 3D. 

Similar to ROST, we model the likelihood of the observed data in terms of the latent topic label variables $z$:
\begin{eqnarray}
\Prob(w| x,t ) = \sum_{k \in K_\text{active}} \Prob(w | z=k) \Prob(z=k | x,t).
\end{eqnarray}

Here the distribution $\Prob(w | z=k)$ models the appearance of the topic label $k$, as is shared across all spatiotemporal locations. The second part of the equation $\Prob(z=k | x,t)$ models the distribution of labels in the spatiotemporal neighborhood of location $(x,t)$. We say that a label is active if there is at least one observation which has been assigned this label. The set of all active labels is $K_\text{active}$.

Let $w_i = v$, be the $i$th observation word with spatial coordinates $x_i$, and time $t_i$, where $i\in[1,N)$, and the observation $v$ is discrete and takes an integer value between $[0,V)$. Each observation $w_i$ is described by latent label variable $z_i=k$, where $k$ again is an integer. 

\begin{eqnarray}
\Prob(w_i=v | z_i=k) =  \frac{n_{v,k} + \beta}{N+V\beta-1}.
\end{eqnarray}

Here $n_{v,k}$ is the number of times an observation of type $v$ has been assigned label $k$ thus far (excluding the $i$th observation), $N$ is the total number of observations, $V$ is the vocabulary size of the observations, and $\beta$ is the Dirichlet parameter for controlling the sparsity of the $\Prob(w|z)$ distribution. A lower value of $\beta$ encourages sparser $\Prob(w|z)$ with peaks on a smaller number of vocabulary words. This encourages topics to describe more specific phenomena, and hence requiring more topics in general to describe the data. A larger value of $\beta$ on the other hand would encourage denser distributions, encouraging a topic to describe more general phenomena in the scene. 

In this work we assume that the set of all distinct observation words is known, and the set has size $V$, however, the number of labels used to describe the data $K$ is inferred automatically from the data. Through the use of Bayesian nonparametric techniques such as Chinese Restaurant Process (CRP), it is possible to model, in a principled way, how new categories are formed\cite{Teh:2006:HDP, Teh2010}. Using CRP, we model whether a word is best explained via an existing label, or by a new, previously unseen label; allowing us to build models that can grow automatically with the growth in the size and complexity of the data.

\begin{eqnarray}
\Prob(z_i=k | z_1,\dots, z_N ) =  
 \begin{cases}
     \frac{n_{k,g_i} + \alpha}{C(i,k)} & k \in K_{\textit{active}} \\
     \frac{\gamma}{C(i,k)} & k = k_{ \text{new}}\\
     0 & \text{otherwise.}
 \end{cases}
\end{eqnarray}

Here $n_{k,g_i} $ is the total number of observations in the spatiotemporal neighborhood of the $i$th observation, excluding itself; dirichlet prior $\alpha$ controls the sparsity of the scene's topic distribution; CRP parameter $\gamma$ controls the growth of the number of topic labels; and  $C(i,k)= \sum_i^N (\Indi_{[n_k>0]}n_{k,g_i}+\alpha)+ \gamma -1$ 
is the normalizing constant.

We use the realtime Uniform+Now Gibbs sampler proposed in \cite{Girdhar2015Gibbs} to compute the posterior topic labels for the datasets. We update the sampler to use the Chinese Restaurant Process for automatic discovery of new labels.

\section{Anomaly Detection}
Given $\Prob(z|t)$, the topic label distribution of a each time step, we can compute the marginal distribution 
\begin{eqnarray}
\Prob(z=k) = \sum_{t=1}^T\frac{ \Prob(z=k|t)}{T},
\end{eqnarray}
which defines the distribution of topic labels for the entire dataset. We can then define the perplexity score $S(t)$ of observations made at a given time-step $t$ as: 
\begin{eqnarray}
S(t) &=& \exp (\Hx(\Prob(z=k|t), \Prob(z=k))\\
     &=& \exp\left(- \sum_k(\Prob(z=k|t)  \log \Prob(z=k)\right). \label{eq:ppx}
\end{eqnarray}

Here the function $\Hx(p,q) = -\sum_x p(x) \log q(x)$ computes the cross entropy of the two distributions $p$ and $q$. If we assume a normal scene has the topic distribution $\Prob(z)$, then $\Hx\left(\Prob(z=k|t), \Prob(z=k)\right)$ computes the average number of bits needed to encode time-step $t$, using codes optimized for distribution is $\Prob(z)$. Taking the exponential of the cross entropy then gives us the uncertainty in assigning topic labels to the given time step.

A time-step where most of the observations were labeled with a commonly occurring topic label will be given a low score, whereas if a time-step with rare topic labels will be given a high score.

\section{Experiments}
To demonstrate the effectiveness of the proposed BNP-ROST scene modeling for anomaly detection, we conducted two experiments on completely different kinds of datasets. The first dataset consists of images collected by a robot as it explores the seafloor, and the second dataset consists of a static camera set in a coral reef, observing a complex and dynamic scene.  

\begin{figure}
\begin{center}
\includegraphics[width=0.7\columnwidth]{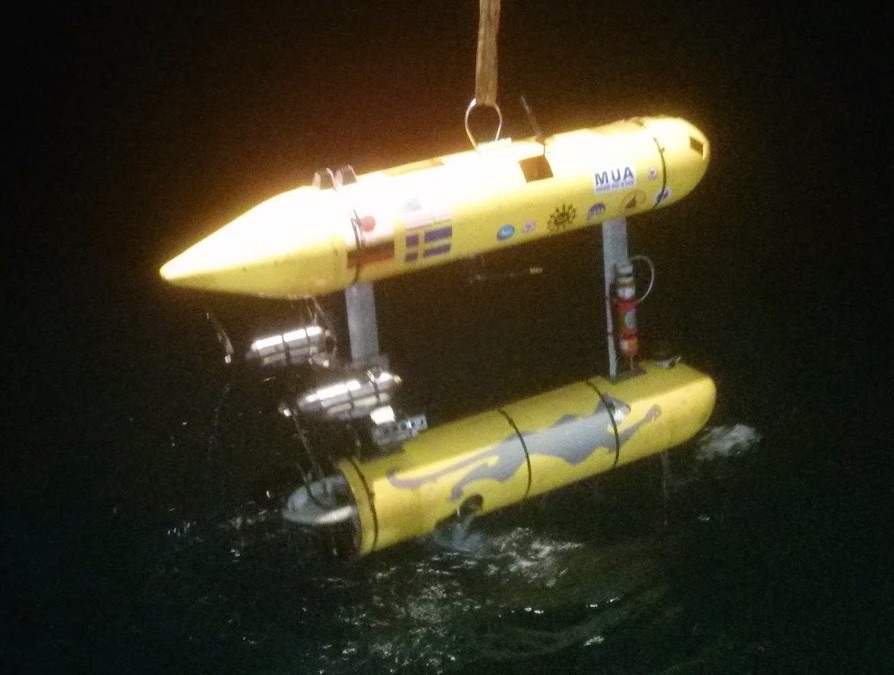}
\caption{Jaguar AUV was used to collect the seafloor image data.}
\end{center}
\end{figure}

\begin{figure*}[t]
\begin{center}
\includegraphics[width=1.8\columnwidth]{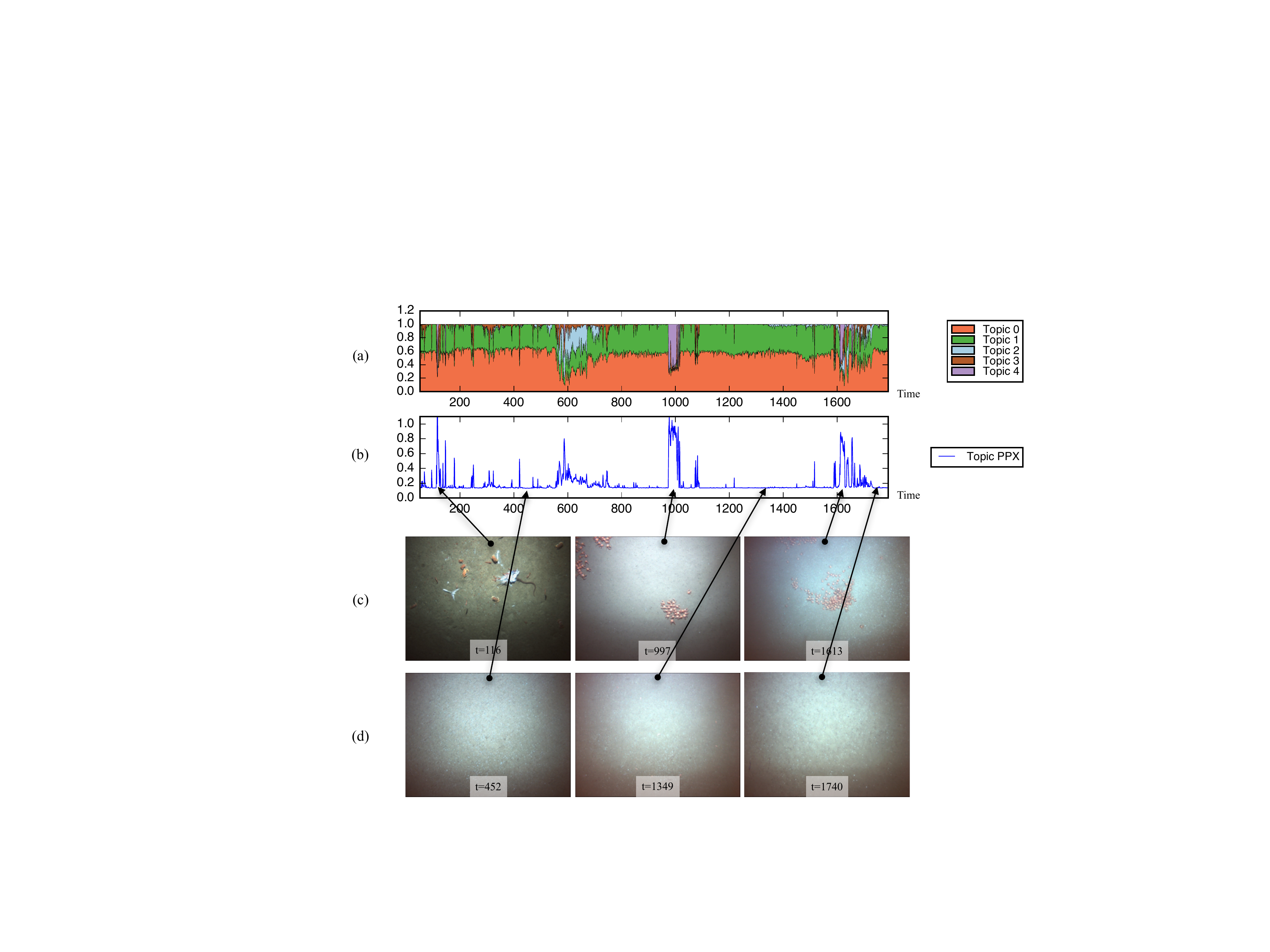}
\caption{Galatheid crab dataset. (a) A stacked plot showing distribution of topic labels at each time step in the dataset, computed using the proposed technique. We see that Topic 0 and 1 are characteristic of the underlying terrain, whereas other topic labels are representative of other phenomena such as galatheid crab aggregations. (b) Shows the normalized perplexity scores for each time step. (c) Shows examples of images with high perplexity scores, corresponding to anomalous observations. Image with $t=116$ shows various animal species feeding off a fish carcass, and is the most anomalous scene in the dataset. Other anomalous observations are of galatheid crabs. (d) Shows examples of some typical images in the dataset, represented by their low perplexity scores.}\label{fig:d19anomaly}
\end{center}
\end{figure*}

\subsection{Fauna Detection in AUV Missions}\label{sec:expcrabs}

In this experiment we analyzed image data collected by the Jaguar AUV \cite{Kunz2009} as it explored the Hannibal seamount in Panama. The mission was conducted primarily at depths beyond 300 meters. At these depths, there is very limited visible fauna. This specific dataset was chosen because it contained observations of galatheid crab gatherings, which is an obviously anomalous phenomena. 

Goal of this experiment was to see if the proposed BNP-ROST algorithm would be able to detect these observations of galatheid crabs, either by characterizing these observations with its own topic label, or by giving them high perplexity scores. Every third image in the dataset was hand-labeled by a team of expert biologists to mark the fauna in the images, which we used as the ground truth for the galatheid crab observations. 

We ran the proposed BNP-ROST algorithm to compute topic distributions for each time step, and the perplexity score $S(t)$ described in Eq. \ref{eq:ppx}. The distribution $\Prob(z|x,t)$ was modeled using cellular approximation described in \cite{Girdhar2015a}, with cell size of 128x128 pixels.  We used $\alpha=0.1, \beta=10, \gamma=1e-5$ for all our experiments. These parameters were chosen after a very sparse grid search in log space of the parameters.

The dataset presented here consists of 1737 images, taken once every three seconds, at an altitude of 4 meters above the seafloor by the Jaguar AUV. The seafloor depth varied between 300-400 meters. 

We extracted the following different kinds of visual words from the data: Textons\cite{Varma2005} with four different orientations, and three different scales, quantized into 1000 different categories using the k-means algorithm; Oriented FAST and rotated BRIEF (ORB) \cite{RubleeE2011} features at FAST detected corners, quantized into 5000 categories; and hue and intensity pixel values distributed on a grid. For each image we extracted 16K texton words, 10K ORB words, and 4K pixel words.

\begin{figure*}[t]
\begin{center}
\includegraphics[width=1.4\columnwidth]{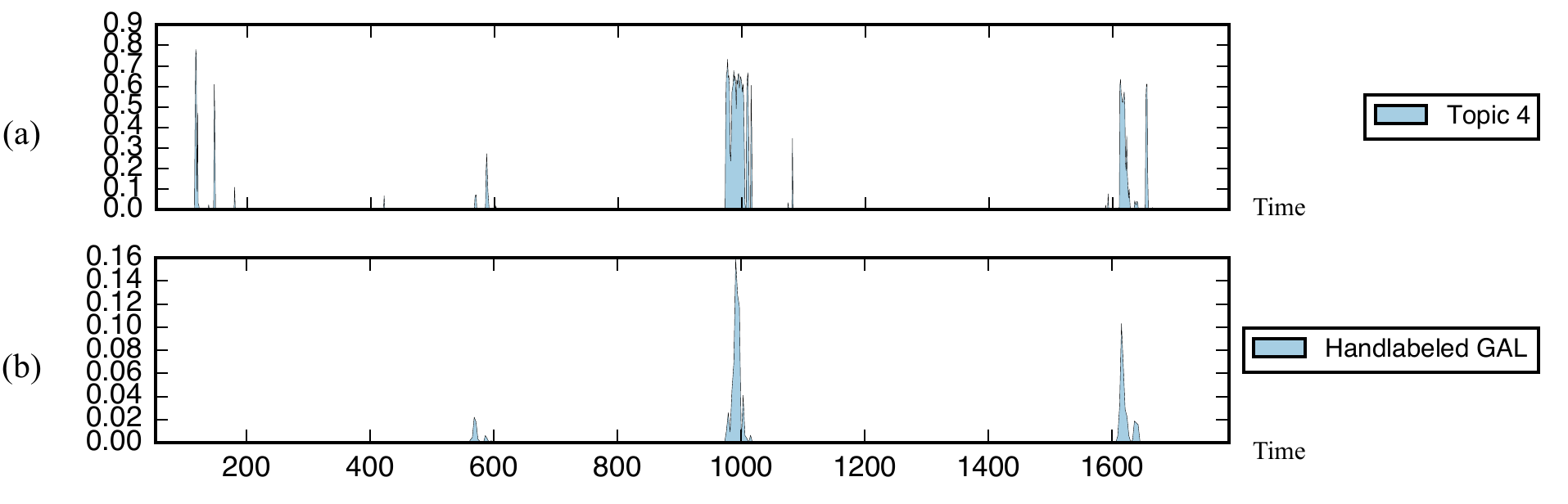}
\caption{Comparison of the learned topic labels with with expert labeled ground-truth for galatheid crab observation. (a) Shows distribution of Topic 4 for each time step, computed automatically by the proposed scene modeling technique, without any supervision. (b) Shows ground normalized distribution of galatheid crabs across the same timeline, labeled by an expert biologist. We see that there are three regions in the AUV mission where the crabs were observed: around time steps 600, 1000, and 1600. All three regions are correctly characterized by the Topic 4 of the scene model, without any supervision.  The peaks around timestep 100 correspond to detection of other kinds of crab (not galatheid), which are labeled with the same topic label as the galatheid crabs by the topic model, but were labeled with a different label by the biologist.}\label{fig:topiccrab}
\end{center}
\end{figure*}

\subsubsection*{Results}

Figure~\ref{fig:d19anomaly}(a) shows the distribution of topic labels over time for the galatheid crab dataset. We see that topic 0 and 1 are representative of the underlying terrain observed by the robot during the mission, and stay consistently represented throughout the timeline, whereas the other topics correspond to more episodic phenomena. 
The plot in Fig.~\ref{fig:d19anomaly}(b) shows the perplexity score of each observation, which was computed given the topic distribution at that time step using Eq.~\ref{eq:ppx}. 

A time-step with low perplexity score implies that it contains images that are characteristic of the entire dataset. Some examples of such images are shown in Fig.~\ref{fig:d19anomaly}(d). These images show the underlying terrain that is represented throughout the mission by topics 0 and 1. Some examples of images with high perplexity scores are shown in  Fig.~\ref{fig:d19anomaly}(c). The highest scoring image corresponds to time step $t=116$, which show a feeding aggregation scene with several different species of crabs, squids and an eel eating a fish carcass. These kinds of feeding aggregations are rare in  the deep ocean, where due to the lack of sunlight there is a limited supply of available nutrients to support life. The other two highest peaks corresponds to observations of the galatheid crab's mating aggregations.  

We found out that distribution of topic 4 over the mission timeline (shown in Fig.~\ref{fig:topiccrab}(a)) matches closely with the distribution of galatheid crabs, annotated manually by a team of expert biologists (shown in Fig.~\ref{fig:topiccrab}(b)).

The Kolmogorov–Smirnov statistic for the topic 4 distribution, given the ground truth crab distribution, found to be $D=0.185$.

The substrate in this experiment is characterized by topics 0 and 1. Our hypothesis on on why two topics represent the underlying terrain and not one is as following. We use a constant Dirichlet parameter $\beta$ to represent the distribution of words for a given topic. This constant parameter implies that all topics are modeled to have word distributions with similar sparseness. Hence, if an entity is ideally represented by a more dense distribution of words, it is likely going to be represented using multiple topics under the current model. This is a limitation of the proposed approach. 

\begin{figure*}[]
\begin{center}
\includegraphics[width=1.62\columnwidth]{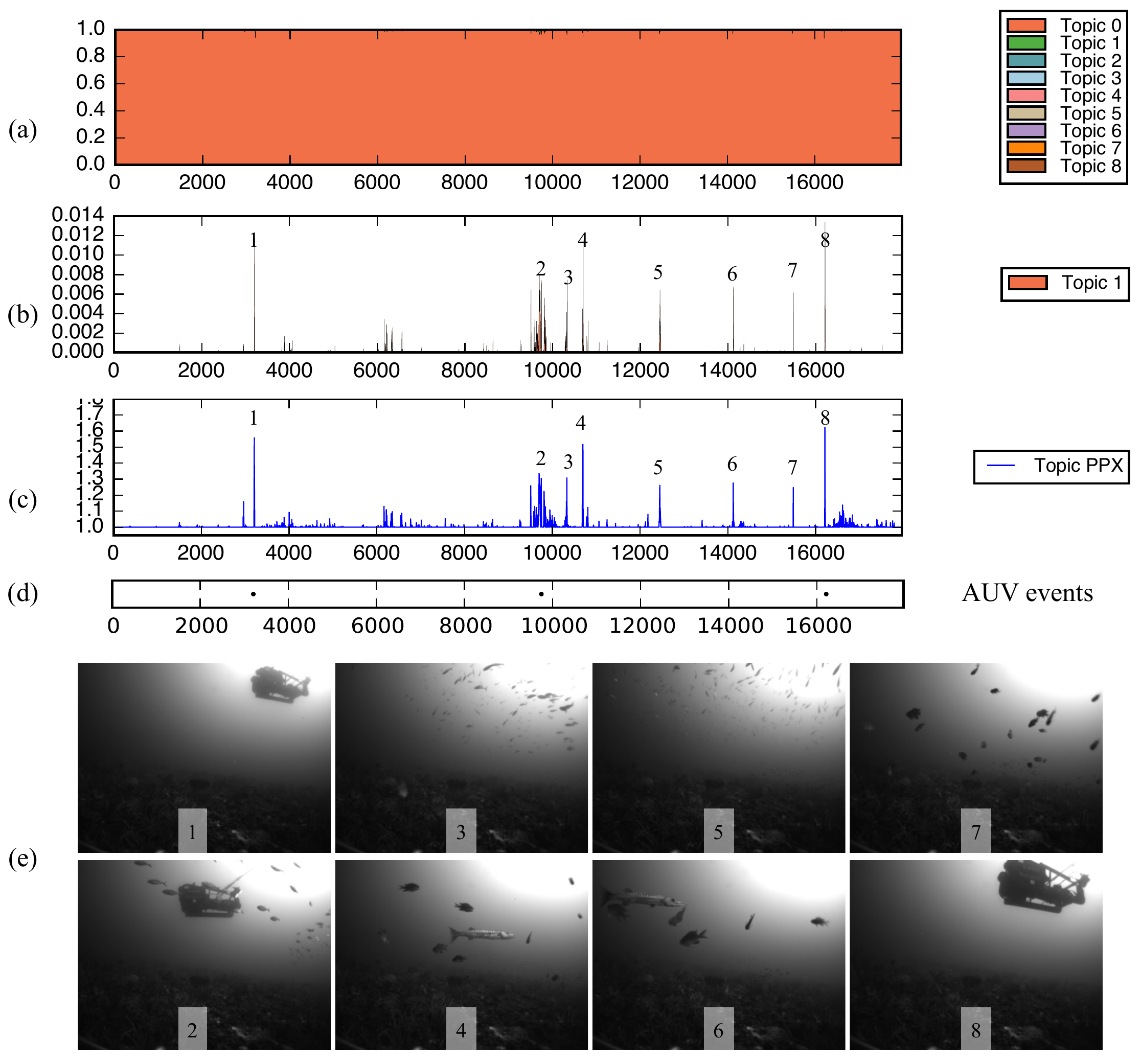}
\caption{Coral reef dataset captured with a static camera. (a) Shows topic label distribution of the dataset across each time step. We see that the distribution is dominated with Topic 0, which characterizes the scene background. (b) Shows the normalized distribution of the next highest weighted topic. (c) Shows the perplexity score for each time step. We see that the perplexity scores correlate well with Topic 1 distribution, which is expected because Topic 1 is relatively rare in the dataset compared to the topic label representing the background. (d) Marks the hand labeled locations of events where the AUV was sighted, which corresponds to peaks 1,2 and 8. (e) Shows images corresponding to the eight highest peaks in the perplexity scores. We see that the perplexity model is able to detect all three events of an underwater vehicle passing in front of the camera. The other perplexity peaks correspond to sightings of barracuda, and schools of fish, both of which are anomalous.}\label{fig:auvdetect}
\end{center}
\end{figure*}

\subsection{Anomaly Detection with a Static Camera in a Complex and Dynamic Scene}
Coral reefs are busy and dynamic environments. As part of another experiment \cite{Campbell2015}, we set up several stationary cameras in the Gulf of Mexico, to characterize the fish behavior in presence of robots. In this experiment we analyzed the image data collected by the stationary seafloor cameras to see if the proposed technique is able to identify an underwater vehicle as anomalous, amongst other observations of fishes, constant flow of debris, lighting change due to wave action on the water surface, and sea plants moving continuously due to current. The dataset consists of an hour long video segment consisting of 17966 image frames, take at the rate of 5 frames per second.  

To focus the topic modeling on the scene foreground, we used a mixture of Gaussian based background detection \cite{Z.Zivkovic2006} technique to compute a background mask for each time step. To characterize the constantly moving  seafloor flora and ocean debris, due to the ocean currents, we use both the intensity values and optical flow values in the background model. We extracted Texton, ORB and intensity words for both the foreground and the background, however the background words were extracted with 1/4$th$ the density of foreground words, to give them less focus. Now to model the scene, we used the same BNP-ROST parameters as described in Sec.\ref{sec:expcrabs}, and computed topic labels for every time step. 

\subsubsection*{Results}
Result of our unsupervised AUV detection experiments are shown in Fig.~\ref{fig:auvdetect}. We see that the topic distribution computed by the proposed technique is dominated by the first topic, which essentially characterizes the scene background. However if we plot the distribution of the second most weighted topic (shown in Fig.\ref{fig:auvdetect}(b)), we see a much more relevant temporal structure of the environment. We find that the peaks of the distribution of Topic 1 match the peak of the perplexity plot of the data (shown in Fig.~\ref{fig:auvdetect}(c)), which can be explained by the fact that Topic 1 is relatively rare (compared to the background Topic 0). In this experiment topic 1 models both the AUVs and the fishes. This is the result of the same problem of constant Dirichlet parameter, which resulted in the two different topic labels being used to characterize the substrate in experiment described in Sec. \ref{sec:expcrabs}. 

The hand labeled AUV sighting events are shown in Fig.~\ref{fig:auvdetect}(d). We see that these events match perfectly with peaks 1,2 and 8.  The other perplexity peaks correspond to sightings of barracudas and large schools of fishes. 

\section{Conclusion}
In this paper we described a Bayesian non-parametric topic modeling technique for modeling semantic content of spatiotemporal data such as video streams, and then used it to identify anomalous observations. We applied the proposed technique to two different kinds of datasets containing observations from unstructured benthic environments. The first dataset containing image data collected by an AUV. We showed that the proposed technique 
was able to automatically identify and characterize observations of galatheid crabs, and the computed distribution matched the hand-labeled distribution with Kolmogorov–Smirnov statistic $D=0.185$. The second data consisted of image data from a  stationary camera set in a busy coral reef, where an underwater robot made three passes in front of the camera. The proposed algorithm was able to identify all three vehicle crossings as anomolous.The fact that the proposed unsupervised algorithm works well in two completely different scenarios, gives us confidence that this approach is well suited for a variety of applications. The Bayesian non-parametric nature of the approach insures that the anomaly models adapts automatically to the data, without requiring careful tuning of the hyper-parameters. Our ongoing efforts are to adapt the proposed technique to be useful with other kinds of data such as audio and sonar imagery. We are also working on using the proposed technique onboard a underwater robot, for context aware adpaptive data collection tasks.

\noindent \textbf{Acknowledgments:}  \small{This work was supported by the Devonshire Foundation, the J. Seward Johnson fund, FQRTN postdoctoral fellowship, and NOAA CINAR.}
\bibliographystyle{IEEEtran_nourl}
\bibliography{IEEEabrv,library}

\end{document}